\documentclass[journal]{IEEEtran} 
\IEEEoverridecommandlockouts
\usepackage{cite}
\usepackage{amsmath,amssymb,amsfonts}
\usepackage{graphicx}
\usepackage{textcomp}
\usepackage{xcolor}
\usepackage{siunitx}
\usepackage{pgfplots}
\usepackage{tikz}
\usetikzlibrary{quantikz}
\usepackage{braket}
\usepackage{tikz-network}
\usepackage{dblfloatfix}
\usepackage[absolute]{textpos}
\usepackage{booktabs}
\usepackage{physics}
\usepackage{braket}
\usepackage[ruled,linesnumbered]{algorithm2e}
\usepackage{algpseudocode}
\usepackage{caption}
\usepackage{subcaption}
\usepackage{bbold}
\usepackage{tikz}

\usepackage{mathptmx}
\usepackage[labelsep=quad,indention=10pt]{subfig}
\usepackage{bm}
\def\BibTeX{{\rm B\kern-.05em{\sc i\kern-.025em b}\kern-.08em
    T\kern-.1667em\lower.7ex\hbox{E}\kern-.125emX}}

\begin{document}
\bstctlcite{IEEEexample:BSTcontrol}
\title{Successive Data Injection in Conditional Quantum GAN Applied to Time Series Anomaly Detection}
%
%
%

\author{Benjamin~Kalfon,
        Soumaya~Cherkaoui,
        Jean-Frédéric~Laprade,
        Ola~Ahmad
        and Shengrui~Wang
\thanks{B. Kalfon and S. Cherkaoui are with Polytechnique Montréal.}
\thanks{JF. Laprade and S. Wang are with Université de Sherbrooke.}
\thanks{O. Ahmad is with Thales Digital Solutions.}}

\maketitle

\begin{abstract}
Classical GAN architectures have shown interesting results for solving anomaly detection problems in general and for time series anomalies in particular, such as those arising in communication networks. In recent years, several quantum GAN architectures have been proposed in the literature.
When detecting anomalies in time series using QGANs, huge challenges arise due to the limited number of qubits compared to the size of the data. To address these challenges, we propose a new high-dimensional encoding approach, named Successive Data Injection (SuDaI). In this approach, we explore a larger portion of the quantum state than that in the conventional angle encoding, the method used predominantly in the literature, through repeated data injections into the quantum state. SuDaI encoding allows us to adapt the QGAN for anomaly detection with network data of a much higher dimensionality than with the existing known QGANs implementations. In addition, SuDaI encoding applies to other types of high-dimensional time series and can be used in contexts beyond anomaly detection and QGANs, opening up therefore multiple fields of application.
\end{abstract}

\begin{IEEEkeywords}
GAN, Anomaly detection, Time series, QGAN, QuGAN, Network anomaly detection, Data encoding, Quantum data encoding
\end{IEEEkeywords}

%
\IEEEpeerreviewmaketitle

\section{Introduction}
%
%
%
%
\IEEEPARstart{T}{ime} series are an omnipresent data type in the scientific and industrial world. Because of their temporal nature, they are widely observed where it is necessary to monitor various variables in systems over time. Examples of the different fields where events occur over time range from finance, to meteorology and networking \cite{tsay_analysis_2005,craddock_analysis_1965,wu_network_2005}.  In all of these fields it is crucial to be able to detect when an abnormal behaviour occurs \cite{botzen_economic_2019,cashell2004economic,doocy2013human}. Thus, anomaly detection in time series has been for decades a very active subjet in data analysis in general \cite{tsay_time_1986}, and in particular in the field of communication and networks. This is all the more important as networks have become an integral part of the infrastructure of today's connected and smart society. Although some anomalies can be mere malfunction or external events, many anomalies in networks come from cyberattacks that are growing in sophistication. As a result, anomaly detection has garnered increasing research attention. Given the substantial consequences of a successful attack, there are strong motivations for attackers to improve their attack methods. Conversely, defenders also have significant incentives to enhance their defense strategies. To gain an edge in this adversarial context,  it can be strongly advantageous to explore innovative technologies that can help tipping the scales in favor of defenders.\\

 In the last decade, classical machine learning (ML) has proven to be highly effective at analyzing and extracting insight from data. As a result, ML has become the primary contributor in the  advancement in the field of anomaly detection, with many of the state-of-the-art propositions using deep learning methods \cite{Pang_2021}. Among these ML techniques, generative methods, more specifically Generative Adversarial Networks or GANs, have shown interesting results\cite{xia2022gan}. For instance, the early work\cite{schlegl_unsupervised_2017} on GANs shows that one can detect anomalies by comparing the outputs of the generative model with real data. Newer approaches extend this concept by adding the utilisation of the encoder/decoder structure found in bidirectional GANs. This enables a more efficient data comparison by employing the latent space output of the encoding process \cite{schlegl_f-anogan_2019,zenati_efficient_2019}. In the context of network time series, because of the large data size, it is common to use a sliding time window to reduce the size of the data to be processed. Models must then take into account the contextual data contained within the window. One way to achieve this is to work on the conditional distribution of future data as a function of previous data. For this purpose, a conditional GAN or cGAN is practical, this type of GAN learns to replicate conditional distributions by taking contextual data  as an input \cite{mirza2014conditional}.  \\

Meanwhile, with the recent progress in quantum hardware \cite{de_leon_materials_2021}, we have started to see experiments that demonstrate a quantum advantage over known classical methods \cite{arute_quantum_2019,kim2023evidence}.
There have been propositions in the field of quantum machine learning drawing inspiration from classical techniques \cite{schuld_introduction_2015,biamonte_quantum_2017,cerezo_variational_2021}. In particular, several quantum GAN (QGAN) architectures have been proposed since the founding article \cite{lloyd_quantum_2018} in 2018 \cite{dallaire-demers_quantum_2018,zoufal_quantum_2019}. Recently, a new kind of QGAN has been developped in several papers: state fidelity based QGANs \cite{stein_qugan_2021,chu_iqgan_2022,niu_entangling_2022}. These offer increased convergence properties when compared to the previous propositions by using a quantum state comparison sub-circuit. \\

Given the demonstrated success of GANs in detecting anomalies and the ongoing progress in QGANs, the prospect of applying QGANs to anomaly detection is becoming an interesting and promising research avenue. However, all the existing QGAN approaches in the literature suffer from the drawback of not putting much emphasis on data encoding. These approaches do not address the specific case of a conditional Quantum Generative Adversarial Network (cQGAN). In fact, when attempting to directly adapt these QGAN architectures for anomaly detection and transform them into cQGANs, we encounter limitations related to the number of qubits. The issue lies in the fact that almost all QGAN architectures described in the current literature rely on angle encoding to represent real data. This encoding requires a number of qubits that scales lineary with the size of the encoded data.  For time series data, where a substantial contextual window of data is used, this encoding quickly becomes impractical.\\

In this paper, we propose a novel approach to tailoring existing QGAN techniques for time series and their high dimensionality. We design a novel scheme, called Successive Data Injection (SuDaI), to allow for an encoding of high dimensional data point into a quantum state sequentially.  We then demonstrate its effectiveness in a quantum design by building an anomaly detection system and testing it on a network anomaly detection dataset that we preprocess into time series data.\\

This article's structure is as follows: section \ref{sec2} provides the necessary background for understanding the proposition by presenting quantum computing basics, QGANs and the constraints and hypothesis for our specific application that lead to the proposition. 
Section \ref{sec3} exposes the design of the proposed quantum encoding method and of the conditional quantum GAN used. Section \ref{sec4} details the functioning of the anomaly detection system built around the QGAN proposition. Section \ref{sec5} presents the details of our implementations including choice of the dataset, experiment description and an analysis of the results of the QGAN and of the anomaly detection system. Section \ref{sec6} summarizes the contributions and the limitations of this work and suggests possible paths for future improvements.

\section{Background}\label{sec2}

\subsection{Quantum Background}

The unit of information of quantum computation \cite{nielsen2001quantum} is called the qubit (for quantum bit). Mathematically, a qubit is a vector in a Hilbert space of dimension 2, $\mathcal{H}_2$. One special orthonormal basis of this vector space is noted as $\{\vec{0},\vec{1}\}$ or more commonly $\{\ket{0},\ket{1}\}$ using Dirac notation. This basis is often named the computational basis and one can then express a qubit $\ket{\varphi}$ in this basis. Let $(\alpha,\beta) \in \mathbb{C}^2$ then,
\begin{align}
\ket{\varphi} = \alpha  \ket{0} + \beta \ket{1}. \label{qubit_def}
\end{align}
Another constraint on qubits is that they should be normalized, meaning $||\varphi||^2 = |\alpha|^2+|\beta|^2 = 1$. This is necessary to define an operation on qubits called a measurement. Measuring a qubit $\ket{\varphi}$ defined as in \eqref{qubit_def} in the computational basis yields either 0 with probability $|\alpha|^2$ or 1 with probability $|\beta|^2$. The normalization ensures that the measurement samples from a proper probability distribution.\\
To manipulate a qubit, any linear transformation that preserves this normalization constraint is allowed. This constitutes the space of unitary operations of $\mathcal{H}_2$. $U$ is said to be unitary if and only if $U U^\dagger = U^\dagger U = I$ where $U^\dagger$ is the conjugate transpose of $U$.\\

These rules are generalized for $n$ qubits by considering the $n^{th}$ power of $\mathcal{H}_2$ with the tensor product, i.e. $\mathcal{H}_2^{\otimes n}$, which is a $2^n$ dimensional Hilbert space. The computational basis in which the quantum states can be expressed (Eq.~\eqref{nqubits}) is then $\{\ket{k}\}_{0\leq k<2^n}$, with $k$ being a binary number. A measure yields one of those states with a probability equal to the square of the corresponding amplitude modulus (Eq.\eqref{nqubits_proba}).  Allowed operations are the unitary operations on  $\mathcal{H}_2^{\otimes n}$.\\

\begin{equation}
    \ket{\varphi} = \sum_{k < 2^n} \varphi_k \ket{k} \label{nqubits}
\end{equation}
\begin{equation}
    P(Measure = k) = |\varphi_k|^2 \label{nqubits_proba}
\end{equation}

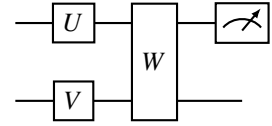
\begin{figure}[h]
    \centering
    \begin{quantikz}
    &\gate{U}&\gate[2]{W}& \meter{} \\
    &\gate{V}&&\qw
    \end{quantikz}
    \caption{An example of a quantum circuit on two qubits with unitaries $U$, $V$, $W$ and a measurement on the first qubit. }
    \label{circ_example}
\end{figure}

From this point, one can manipulate a system of qubits (also called a quantum state) by applying successive unitary operations to it. The common representation of quantum algorithms is a quantum circuit (see Fig.~\ref{circ_example}).\\

In this example circuit, each line represents a single qubit, a box is a unitary (or quantum gate in this context) that is applied to all the qubits that pass through the box. Following this convention, $W$ is a two-qubits gate. When two unitaries $U$ and $V$ are applied in parallel to different qubits, it is equivalent to the unitary $U \otimes V$ applied on the whole quantum state.

\subsection{GANs and Quantum GANs}

\begin{figure}[htbp]
    \centering
    \includegraphics[scale=0.5]{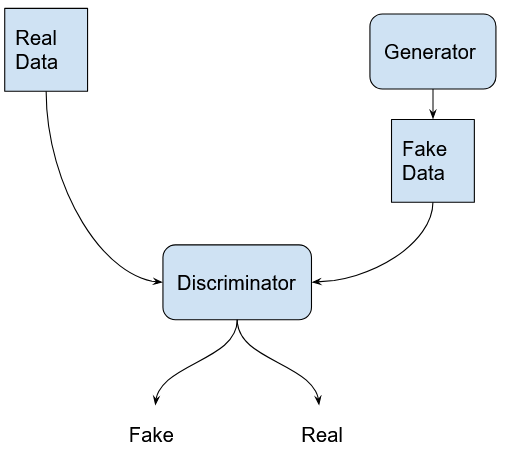}
    \caption{A schematic of the pipeline of a GAN.}
    \label{GAN}
\end{figure}

Generative adversarial networks consist of 2 agents: the discriminator and the generator \cite{goodfellow_generative_2014}. The discriminator is a classifier trained on a given real distribution. The generator produces fake data with the goal of fooling the discriminator into classifying it as part of the real distribution it learns to recognize. The discriminator is thus also trained to recognize the data of the generator as not being part of the real distribution. At the end of this adversarial game, the generator should generate data according to the real distribution. A schematic describing the pipeline of a GAN is shown in Fig. \ref{GAN}.\\

These discriminator and generator are typically implemented as neural networks and are optimized through gradient descent methods. Analogous agents in quantum computing have already been proposed in the literature, such as Quantum Variational Circuits (QVCs) \cite{biamonte_quantum_2017,cerezo_variational_2021}. They consist of quantum circuits with parametric gates, the parameters are then optimized with classical techniques using gradient descent with respect to an objective function. To be able to perform such optimization, one needs to access the derivatives of the circuit, which are obtainable through the parameter shift rule \cite{wierichs_general_2021,crooks_gradients_2019}.\\

In this work, we will be using $\boldsymbol{\theta}$ to represent the vector containing all the parameters used in a given circuit, $\theta$ or $\theta_j$ will be used to represent a single real parameter.
Let $l$ be a loss function for our quantum variational circuit that depends on a real parameter $\theta_i \in \boldsymbol{\theta}$, then
\begin{equation}
    \dv {l}{\theta_i}(\boldsymbole{\theta}) = \frac{1}{2} \left(l\left(\boldsymbol{\theta} + \frac{\pi}{2}\mathbb{1}_i\right) - l\left(\boldsymbol{\theta} - \frac{\pi}{2}\mathbb{1}_i\right)\right), \label{eq_parshift}
\end{equation}

\noindent where $\mathbb{1}_i$ is a vector that has the same number of dimensions as $\boldsymbol{\theta}$ and takes value 1 at index $i$ and 0 otherwise.\\

\begin{figure}[h]
     \centering
     \begin{subfigure}[b]{0.5\textwidth}
         \centering
          \begin{quantikz}
    \lstick{$\ket{0}$}&\phantomgate{}&\gate[3]{\textsc{SWAP TEST}}& \meter{} \\
    \lstick{$\ket{0}^{\otimes n}$}&\gate{D_{\boldsymbol{\theta} }}&&\qw\\
    \lstick{$\ket{x}$}&\phantomgate{}&&\qw
    \end{quantikz}
    \caption{The discriminator $D_{\boldsymbol{\theta}}$ and the real data $\ket{x}$.}
    \label{DnD_sfQGAN}
     \end{subfigure}
     \hfill
     \vspace{5mm}
     \begin{subfigure}[b]{0.5\textwidth}
         \centering
         \begin{quantikz}
    \lstick{$\ket{0}$}&\qw &\gate[3]{\textsc{SWAP TEST}}& \meter{} \\
    \lstick{$\ket{0}^{\otimes n}$}&\gate{D_{\boldsymbol{\theta}}}&&\qw\\
    \lstick{$\ket{0}^{\otimes m}$}&\gate{G_{\boldsymbol{\theta}}}&&\qw
    \end{quantikz}    \caption{The discriminator $D_{\boldsymbol{\theta}}$ and the generator $G_{\boldsymbol{\theta}}$.}
    \label{DnG_sfQGAN}
     \end{subfigure}
        \caption{State fidelity QGAN circuit designs.}
        \label{fig:sfQGAN}
\end{figure}
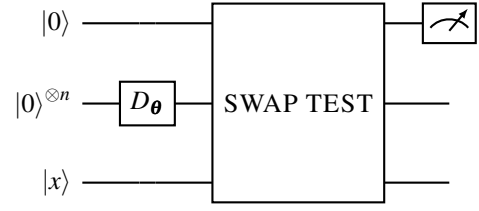
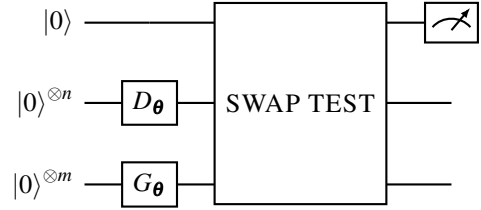

The type of quantum GAN that will be used in this work is one that has been recently developed, namely state fidelity quantum GANs \cite{stein_qugan_2021,chu_iqgan_2022,niu_entangling_2022}. The QGAN is described by the two circuits shown in Fig.\ref{DnD_sfQGAN} and Fig.\ref{DnG_sfQGAN}. The $D_{\boldsymbol{\theta}}$ and $G_{\boldsymbol{\theta}}$ unitaries are quantum variational circuits for the discriminator and the generator. $\ket{x}$ is an encoding of a real data point $x$ inside a quantum register. \\

\begin{figure}[htbp]
    \centering
  \begin{quantikz}
   \lstick{$\ket{0}$} &\gate{H}\gategroup[3,steps = 3]{SWAP test} 
        &\ctrl{2} & \gate{H} &\meter{}\\
    \lstick{$\ket{\varphi}$} & \qw &\targX{} & \qw  &\qw\\
    \lstick{$\ket{\psi}$} & \qw &\swap{-1} &\qw   &\qw
    \end{quantikz}    \caption{Circuit of the SWAP test composed of two Hadamard gates and a controlled SWAP gate.}
    \label{SWAP}
\end{figure}
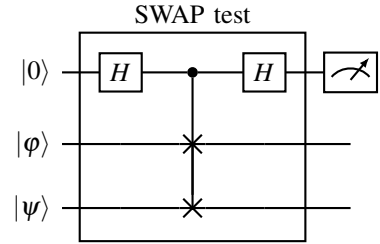

The component at the end of the two circuits is called the SWAP test, which takes as input two quantum states $\ket{\varphi}$ and $\ket{\psi}$ with the same dimension, and outputs on an auxiliary qubit, a state that encodes the similarity between the two input quantum states \cite{stein_qugan_2021}. The details of this part of the circuits can be seen in Fig.\ref{SWAP}. When measuring this qubit the output follows the probability given by eq.~\eqref{eq_fidelity}. This value is called the fidelity between the states $\ket{\varphi}$ and  $\ket{\psi}$. It can be estimated by sampling from the auxiliary qubit.

\begin{equation}
P(Measure = 0 ) = \frac{1}{2}(1 + \braket{\varphi|\psi}^2 ), \label{eq_fidelity}
\end{equation}

\noindent Finally, the similarity $\braket{\varphi|\psi}$, which is the Hermitian product between $\ket{\varphi}$ and $\ket{\psi}$, can be extracted and used to construct a loss for the generator and discriminator.

The SWAP test is used inside the QGAN design to compare the output states of the generator and discriminator circuit or the discriminator's output state with the real data representation. The controlled SWAP operation requires both input states to have the same number of qubits. In practice we designate sub-registers for the discriminator and the generator. These registers must also have a number of qubits equal to of the state representing the real data. The remaining qubits outside the registers are not used. This means that the generator and discriminator have a number of qubits greater than or equal to that of the state encoding real data.
In most cases reported in the literature, all the qubits of the generator and discriminator are used and thus both circuits use the same number of qubits. This number is also the dimension of the data encoded with a basic angle encoding. In our work, as the generated point is one dimensional, the sub registers will be both composed of one qubit.\\

Note that in this QGAN design, the generator and discriminator have very similar architectures. In fact, in most published cases, the discriminator and generator sub-circuits employs the same architecture. They are distinguished by the different roles played inside the adversarial game. Those roles are determined by their attributed loss functions. For our purpose, they will be respectively: i) $L_{D,X}$, the discriminator loss against real data $x$ (eq.~\eqref{eq_loss_D_data}), ii) $L_ {D,G}$, the discriminator loss against the generator (eq.~\eqref{eq_loss_D_gen} and iii) $L_{G,D}$, the generator loss against the discriminator (eq.~\eqref{eq_loss_G}):
\begin{align}
L_{D,X}(x)  &=  -log( E[\braket{\varphi|x}^2]) \label{eq_loss_D_data} \\
L_ {D,G}  &=  -log(1 - E[\braket{\varphi|\psi}^2]) \label{eq_loss_D_gen}\\
L_{G,D}  &=  -log(E[\braket{\varphi|\psi}^2]) \label{eq_loss_G}
\end{align}

\subsection{Limitations in Quantum Data Loading}

Recall that we aim to detect anomalies from windowed time series data using a quantum generative model. This type of data is typically high dimensional depending on the number of time steps that may be in the thousands or more. This is challenging given the current constraints on quantum hardware, which is for now limited in terms of number of qubits, connectivity between qubits and coherence times. Even though a quantum state of $n$ qubits could encode an exponential amount of data in its amplitudes, preparing this quantum state has a time and space complexity that makes this approach not viable. The most common data encoding strategy used in quantum variational algorithms is known as angle encoding where $n$ real values are encoded into $O(n)$ qubits by applying a parametrized rotation gate to each qubit. This strategy has the advantage of requiring few computational resources when compared to more information dense methods such as amplitude encoding. However, angle encoding requires a linear number of qubits and does not take advantage of the exponentially large vector space.

\section{The Proposed Quantum GAN}\label{sec3}

\subsection{The Conditional Quantum GAN}
\begin{figure}[htbp]
    \centering
    \includegraphics[scale=0.5]{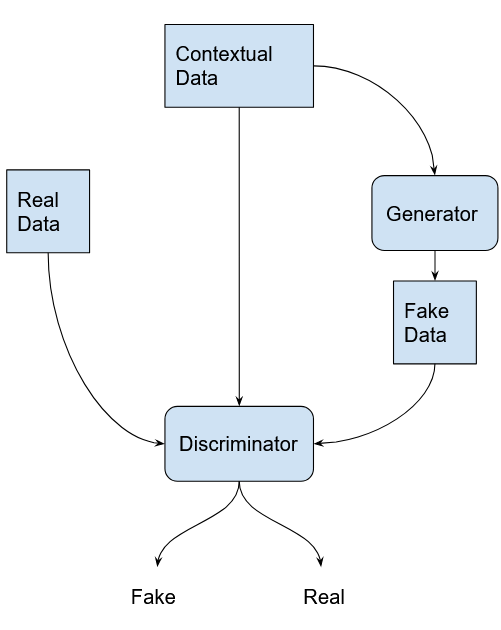}
    \caption{Pipeline of a conditional GAN.}
    \label{cGAN}
\end{figure}

To reduce the dimensionality of the data, our approach involves employing a well-established technique commonly used in anomaly detection, consisting in the utilization of a shifting time-window\cite{blazquez2021review}. The core concept behind this method is to identify anomalies within a time series at a specific time step by considering the preceding data points within a defined time window.
Instead of dealing with time series data with an unlimited number of time steps, we simplify the problem by working with fixed-length time windows. Given that our problem now encompasses a conditional distribution, we build a slightly modified architecture for the QGAN, namely a conditional QGAN or a cQGAN. Specifically, both the generator and discriminator need to be preceded by an encoder circuit, which encodes the data from the preceding time steps of the time series. In a more general description, we want to learn the probability distribution of B given A. Fig.\ref{cGAN} describes the pipeline of a conditional GAN \cite{mirza2014conditional}.\\

\begin{figure}
     \centering
     \begin{subfigure}[b]{0.5\textwidth}
         \centering
          \begin{quantikz}
    \lstick{$\ket{0}$}&\qw &\qw  &\gate[3]{\textsc{SWAP test}}  &\meter{} \\
    \lstick{$\ket{0}^{\otimes n}$} &\gate{E_{\boldsymbol{\theta}}(a)}\gategroup[1, steps = 2]{$D_{\boldsymbol{\theta}}$}  &\gate{F_{\boldsymbol{\theta}}}  &  &\qw\\
    \lstick{$\ket{b}$}&\qw &\qw  &  &\qw
    \end{quantikz}
    
    \caption{The discriminator and the real target data $\ket{b}$.}
    \label{cqGAN_DnD}
     \end{subfigure}
     \hfill
     \vspace{5mm}
     \begin{subfigure}[b]{0.5\textwidth}
         
         \begin{quantikz}
    \lstick{$\ket{0}$}&\qw&\qw&\gate[3]{\textsc{SWAP test}}& \meter{} \\
    \lstick{$\ket{0}^{\otimes n}$}&\gate{E_{\boldsymbol{\theta}}(a)}\gategroup[1, steps = 2]{$D_{\boldsymbol{\theta}}$}&\gate{F_{\boldsymbol{\theta}}}&&\qw\\[0.6cm]
    \lstick{$\ket{0}^{\otimes n}$}&\gate{E_{\boldsymbol{\theta}}'(a)}\gategroup[1, steps = 2]{$G_{\boldsymbol{\theta}}$}&\gate{F_{\boldsymbol{\theta}}'}&&\qw
    \end{quantikz}    
    \caption{The discriminator and the generator. }
    \label{cqGAN_DnG}
    \end{subfigure}
        \caption{State fidelity conditional QGAN circuit design. }
        \label{cqGAN}
\end{figure}
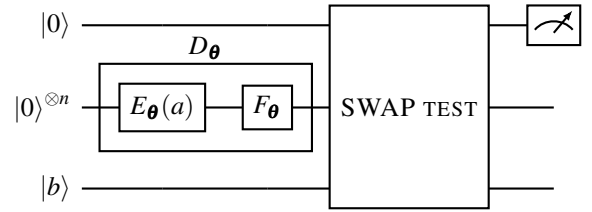
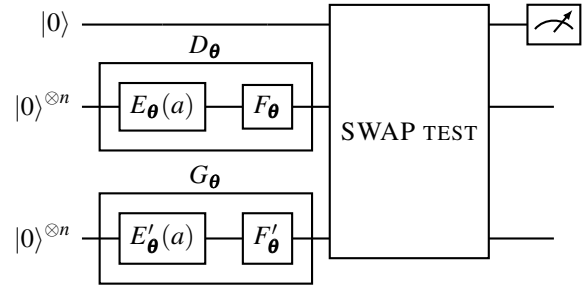

The main difference with typical GANs is that both the generator and the discriminator are fed with the contextual data $a$ in order to discriminate or generate the $b$ object. In the case of a sliding time window in a time series, $a$ is the data contained in the time window and $b$ is the next data point. The architecture for the cQGAN then becomes the one shown in Fig.\ref{cqGAN_DnD} and Fig.\ref{cqGAN_DnG}. In this architecture, both the discriminator and the generator circuits are built in two parts: an encoder ($E_{\boldsymbol{\theta}}$ or $E_{\boldsymbol{\theta}}'$) that takes the contextual data $a$ and creates a quantum state that is then processed by another quantum variational circuit ($F_{\boldsymbol{\theta}}$ or $F_{\boldsymbol{\theta}}'$).\\

To train the our cQGAN, we apply the basic routine described in Algorithm \ref{training}. Similar to training a classic GAN, this training routine is decomposed into three phases. For a certain number of epochs $e$ (line \ref{line_epochs}), we train successively the discriminator on real data (lines \ref{line_DX_start}-\ref{line_DX_end}), the generator on the discriminator (lines  \ref{line_DG_start}-\ref{line_DG_end}), and the discriminator on the generator (lines  \ref{line_GD_start}-\ref{line_GD_end}). Each of these steps is repeated a certain number of times designated by the respective training counts $C_{D,X}, C_{D,G}, C_{G,D}$. These training counts are hyperparameters and require careful tuning to strike a suitable balance between the convergence of the discriminator and the generator.
To train one of the agents we sample a subset of couples $(a,b)$ of contextual data and target data (lines $8,13,18$). For each couple, we feed the agents with the contextual data and compute the derivatives of the appropriate loss function with respect to the parameter vector ${\boldsymbol{\theta}}$ (lines \ref{line_grad_DX},\ref{line_grad_DG},\ref{line_grad_GD}). Then we update ${\boldsymbol{\theta}}$ by taking a gradient step with an amplitude controlled by the learning rate $\alpha$ (lines \ref{line_step_DX},\ref{line_step_DG},\ref{line_step_GD}).

\begin{algorithm}
\SetKw{Init}{Initialize:\\}
\SetKw{Sample}{Sample}
\caption{Training of the conditional quantum GAN }\label{training}
\Init{ Learning rate $\alpha$ \\
    \qquad Circuit vector parameter ${\boldsymbol{\theta}}$ \\
    \qquad Dataset $X = (A,B)$\\
    \qquad Number of epochs $e$\\
    \qquad Training counts $C_{D,X}, C_{D,G}, C_{G,D}$}\\
\For{$\xi \leftarrow 1$ \KwTo $e$}{ \label{line_epochs}
\tcc{Training Discriminator on Real Data: ${D,X}$}
\Sample{$X_{D,X}$ from $X$ with $|X_{D,X}|=C_{D,X}$} \\ \label{line_DX_start}
\For{$x=(a,b) \in X_{D,X}$}{
    $\nabla_{\boldsymbol{\theta}} = \frac{dL_{D,X}(a,b)}{d{\boldsymbol{\theta}}}$\\ \label{line_grad_DX}
    ${\boldsymbol{\theta}} = {\boldsymbol{\theta}} - \alpha$  $\nabla_{\boldsymbol{{\boldsymbol{\theta}}}}$ \label{line_step_DX}
    
}\label{line_DX_end}
\tcc{Training Generator on Discriminator: ${D,G}$}
\Sample{$X_{D,G}$ from $X$ with $|X_{D,G}|=C_{D,G}$}\\ \label{line_DG_start}
\For{$x=(a,b) \in X_{D,G}$}{
    $\nabla_{\boldsymbol{\theta}} = \frac{dL_{D,G}(a)}{d{\boldsymbol{\theta}}}$\\\label{line_grad_DG}
    ${\boldsymbol{\theta}} = {\boldsymbol{\theta}} - \alpha$  $\nabla_{\boldsymbol{\theta}}$ \label{line_step_DG}
    
}\label{line_DG_end}
\tcc{Training Discriminator on Generator: ${G,D}$}
\Sample{$X_{G,D}$ from $X$ with $|X_{G,D}|=C_{G,D}$}\\ \label{line_GD_start}
\For{$x=(a,b) \in X_{G,D}$}{
    $\nabla_{\boldsymbol{\theta}} = \frac{dL_{G,D}(a)}{d{\boldsymbol{\theta}}}$\\\label{line_grad_GD}
    ${\boldsymbol{\theta}} = {\boldsymbol{\theta}} - \alpha$  $\nabla_{\boldsymbol{\theta}}$\label{line_step_GD}
}\label{line_GD_end}
}
\end{algorithm}

\subsection{SuDaI: Encoding through Successive Data Injection}

\begin{figure}[htbp]
    \centering
    \begin{quantikz}
        \lstick[2]{$\ket{0}^n$}& \gate[2]{I_{\boldsymbol{\theta}}^{(1)}(a_1)}&\gate[2]{V_{\boldsymbol{\theta}}^{(1)}}& \qw \ldots & \gate[2]{I_{\boldsymbol{\theta}}^{(K)}(a_K)}&\gate[2]{V_{\boldsymbol{\theta}}^{(K)}}&\qw \\[2cm]
        &&& \qw \ldots & &&\qw
    \end{quantikz}
    \caption{Architecture of the successive data injection circuit with K input and encodinglayers.}
    \label{seq_data_inj}
\end{figure}
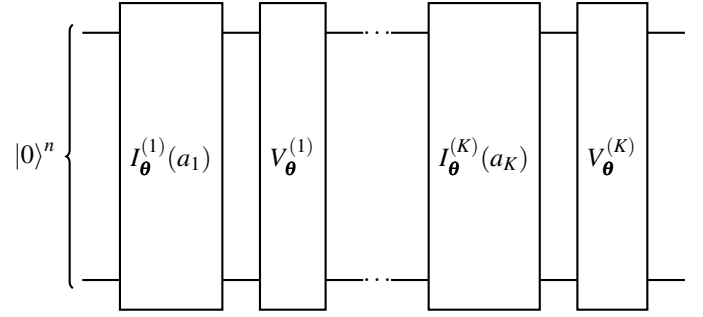

With this architecture still comes the need to encode data larger than the number of qubits. To do so we introduce a new scheme called successive data injection. The principle is to incorporate data into the quantum state in order to, step by step, build a representation that encodes all the useful data for the task.
This results in an encoding method that can accommodate a greater number of data points per number of available qubits, compared to what can be achieved using a basic angle encoding approach.
The design consists of a stack of input layers which depend on the data and trainable parameters and variational layers which depend on trainable parameters only (see Fig. \ref{seq_data_inj}). This approach is inspired by previous works on quantum data reloading \cite{schuld2021effect} and most notably \cite{perez2020data}  where the authors propose to re-upload data points at different points of the quantum variational circuit to improve performance. In this work we take it a step further by applying this to sequential data and uploading different data points at each layer.\\

This scheme consists of a succession of input layers $I_{\boldsymbol{\theta}}^{(k)}(a_k)$ and variational layers $V_{\boldsymbol{\theta}}^{(k)}$ for $K$ data points $(a_1,...,a_K)$ in $a$.
The $k^{th}$ input layer consists of a circuit that takes into account both the data point $a_k$ and the parameter vector ${\boldsymbol{\theta}}$. The role of this layer is to inject data into the quantum state in the most general manner possible.  possible
The variational layer only depends on ${\boldsymbol{\theta}}$ and aims at preparing the quantum state for the next injection.\\

\begin{figure}[h]
    \centering
    \begin{quantikz}
        &\gate{R_X(\theta_{k_1})}\gategroup[4,steps = 5]{$V_{\boldsymbol{\theta}}^{(k)}$} 
        &\gate{R_Z(\theta_{k_2})} & \ctrl{1} &\qw &\qw&\qw\\
        & \gate{R_X(\theta_{k_3})} &\gate{R_Z(\theta_{k_4})} & \targ{} & \ctrl{1} &\qw&\qw\\
        & \gate{R_X(\theta_{k_5})} &\gate{R_Z(\theta_{k_6})} &\qw &\targ{} & \ctrl{1} &\qw\\
        & \gate{R_X(\theta_{k_7})} &\gate{R_Z(\theta_{k_8})} &\qw &\qw &\targ{} &\qw 
    \end{quantikz}
    \caption{Variational circuit design.}
    \label{var_layer}
\end{figure}
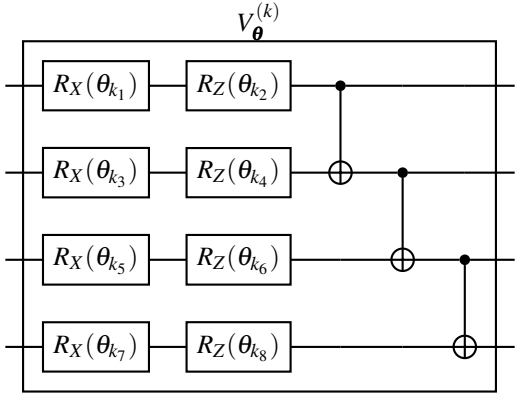

In theory, the circuit design of a such layer can take various forms. Numerous designs have been proposed and studied in the literature, particularly in \cite{sim_expressibility_2019}. In this work, the authors benchmark different variational circuit designs by using two metrics. The first one is the Kullback-Leibler divergence for expressibility of a circuit, i.e. the ability of the circuit to explore most of the Hilbert Space. The second metric is the entangling capability which measures the ability of a circuit to generate entangled states.
The design of the variational layer $V_{\boldsymbol{\theta}}^{(k)}$ used in this work is shown in Fig. \ref{var_layer} for 4 qubits. This design has been chosen among  several ones described in \cite{sim_expressibility_2019}, because of the good performance it exhibits in both metrics when used repeatedly while requiring a relatively low gate count per layer.\\

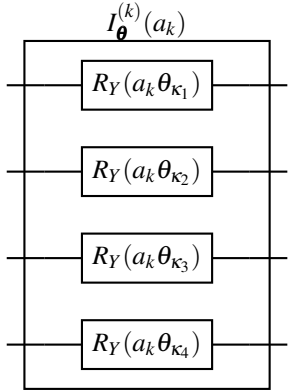
\begin{figure}[htbp]
    \centering
    \begin{quantikz}
        &\qw \gategroup[4,steps = 3]{$I_{\boldsymbol{\theta}}^{(k)}(a_k)$} &\gate{R_Y(a_k \theta_{\kappa_1})}&\qw &\qw\\
        &\qw &\gate{R_Y(a_k \theta_{\kappa_2})} &\qw  &\qw\\
        &\qw &\gate{R_Y(a_k \theta_{\kappa_3})} &\qw  &\qw\\
        &\qw &\gate{R_Y(a_k \theta_{\kappa_4})} &\qw  &\qw
    \end{quantikz}
    \caption{Design of the input layer sub-circuit.}
    \label{input_layer}
\end{figure}

The input layer specific design is described in Fig. \ref{input_layer} for four qubits. For each qubit, two parameters are used to specify the rotation angle of the $R_Y$ gate: one to load the data and the other one as a weight that is learned during the training process. More precisely, the same data point $a_k$ is encoded on each qubit but with different variational parameters. The repetition of such encoding has shown interesting result to increase the functionality of quantum machine learning applications \cite{schuld2021effect}. This creates a very general encoding procedure as all the qubits can be subject to rotations of different intensities. 

\begin{figure*}[h]
\centering
    \begin{tikzpicture}
    \node[scale=0.8]{
    \begin{quantikz}[]
        \lstick{$\ket{0}$}
        &\qw &\qw
        &\qw &\qw &\qw &\qw &\qw 
        &\qw &\qw
        &\qw &\qw &\qw &\qw &\qw 
        & \gate{H}\gategroup[5,steps = 3]{SWAP TEST} &\ctrl{4} &\gate{H} 
        &\meter{}
        \\
        \lstick{$\ket{0}$}
        &\gate{R_Y(a_1 \theta_1)}\gategroup[3,steps = 1]{$I_{\boldsymbol{\theta}}^{(1)}$} &\qw
        &\gate{R_X(\theta_4)}\gategroup[3,steps = 4]{$V_{\boldsymbol{\theta}}^{(1)}$} &\gate{R_Z(\theta_7)} & \ctrl{1} &\qw &\qw
        &\gate{R_Y(a_2 \theta_{10})}\gategroup[3,steps = 1]{$I_{\boldsymbol{\theta}}^{(2)}$} &\qw
        &\gate{R_X(\theta_{13})}\gategroup[3,steps = 4]{$V_{\boldsymbol{\theta}}^{(2)}$} &\gate{R_Z(\theta_{16})} & \ctrl{1} &\qw &\qw
        &\qw &\qw &\qw
        &\qw
        \\
        \lstick{$\ket{0}$}
        &\gate{R_Y(a_1 \theta_2)} &\qw 
        & \gate{R_X(\theta_5)} &\gate{R_Z(\theta_8)} & \targ{} & \ctrl{1} &\qw
        &\gate{R_Y(a_2 \theta_{11})}&\qw
        & \gate{R_X(\theta_{14})} &\gate{R_Z(\theta_{17})} & \targ{} & \ctrl{1} &\qw
        &\qw &\qw &\qw
        &\qw
        \\
        \lstick{$\ket{0}$}
        &\gate{R_Y(a_1 \theta_3)} &\qw
        & \gate{R_X(\theta_6)} &\gate{R_Z(\theta_9)} &\qw &\targ{} &\qw
        &\gate{R_Y(a_2 \theta_{12})}&\qw
        & \gate{R_X(\theta_{15})} &\gate{R_Z(\theta_{18})} &\qw &\targ{} &\qw
        &\qw &\targX{} &\qw
        &\qw
        \\
        \lstick{$\ket{0}$}
        &\gate{R_Y(b)} &\qw
        &\qw &\qw &\qw &\qw &\qw 
        &\qw &\qw
        &\qw &\qw &\qw &\qw &\qw 
        &\qw &\swap{-1} &\qw
        &\qw
    \end{quantikz} 
    };
    \end{tikzpicture}
    \caption{Example of a full circuit discriminator with two contextual data points, two qubits for the discriminator circuit and one qubit for the real data representation, the swap test is performed between the fourth and fifth qubits.} 
    \label{full_circuit}
\end{figure*}
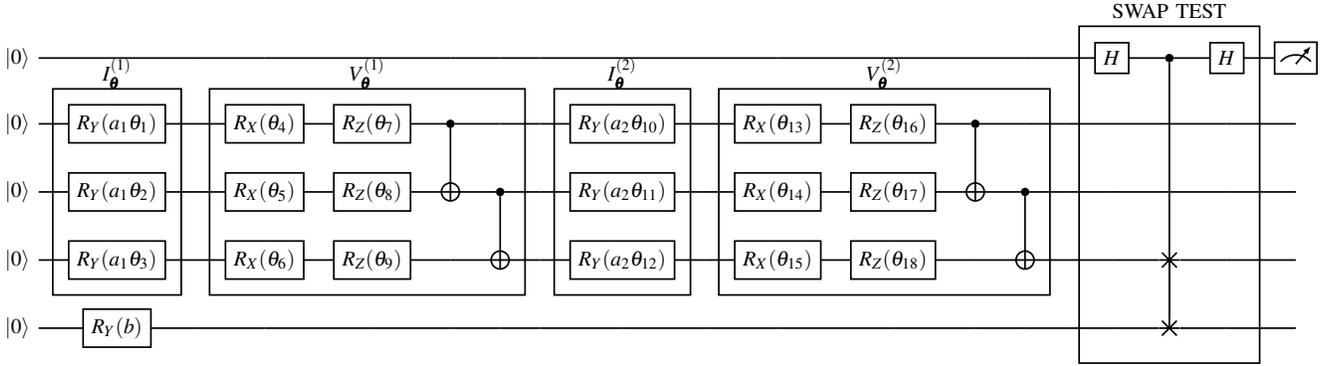

It is also worth noting that when combining the input layer with the variational layer, three consecutive rotations $R_Y$, $R_X$ and  $R_Z$ are applied on each qubit. These three rotations spanning the three axis are sufficient for expressing any one qubit unitary transformation \cite{nielsen2001quantum}.
An example of the complete circuit can be seen in Fig.\ref{full_circuit}.

\section{Anomaly Detection}\label{sec4}

\subsection{DataSet}

\begin{figure*}[t]
    \centering
    \includegraphics[scale= 0.4]{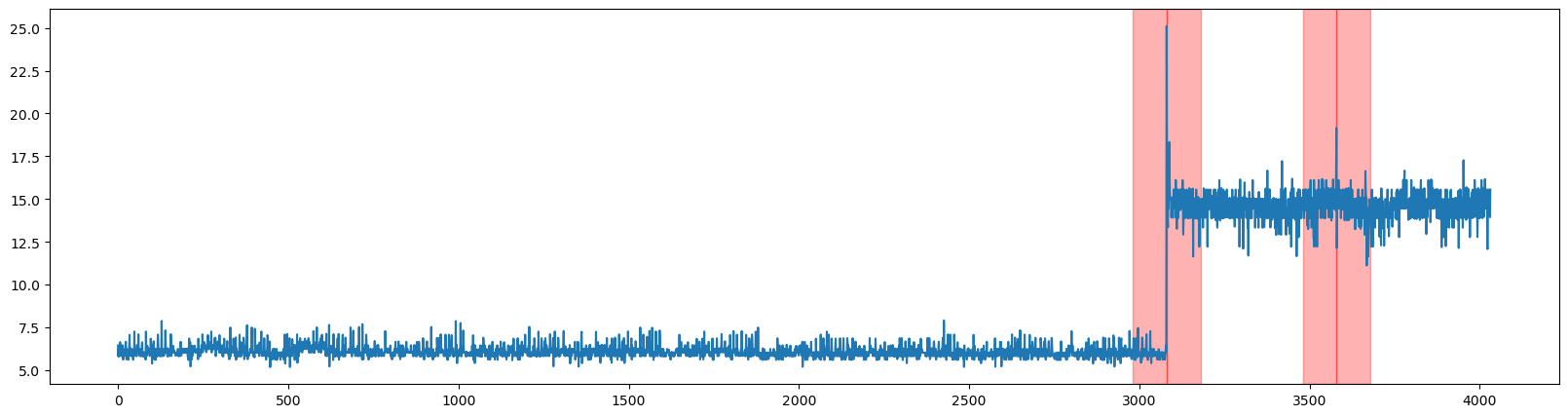}
    \caption{An example of a NAB time series monitoring CPU usage on a radio data system with two anomalies among 4000 time-steps. The anomalies are marked by the red lines while their corresponding time-windows for correct flagging are represented by the pale red areas.}
    \label{example_time_series}
\end{figure*}

To apply our cQGAN to anomaly detection for time series we have used the Numenta Anomaly Benchmark \cite{ahmad2017unsupervised} (NAB) database. Specifically, we chose real world data coming from the sub-dataset called AWSCloudWatch. The time series consists of key performance indicators (KPI) recorded on servers during operation such as CPU usage. Anomalies in the server environment are then translated to anomalies in the time series of records of the different KPIs. The time series are univariate and the time step at which anomalies happen are labeled. The NAB also provides time windows around the anomalies that correspond to the time intervals inside which anomalies flagged by a model should be considered as true positives. A good anomaly detection system should therefore detect anomalies in those windows. One example of a time series with an anomaly and its associated time window can be seen in Fig. \ref{example_time_series}. The time windows are the areas in pale-red and the vertical red lines correspond to the timestamp with an anomalous data point.\\

\begin{figure*}[t]
    \centering
    \includegraphics[scale = 0.4]{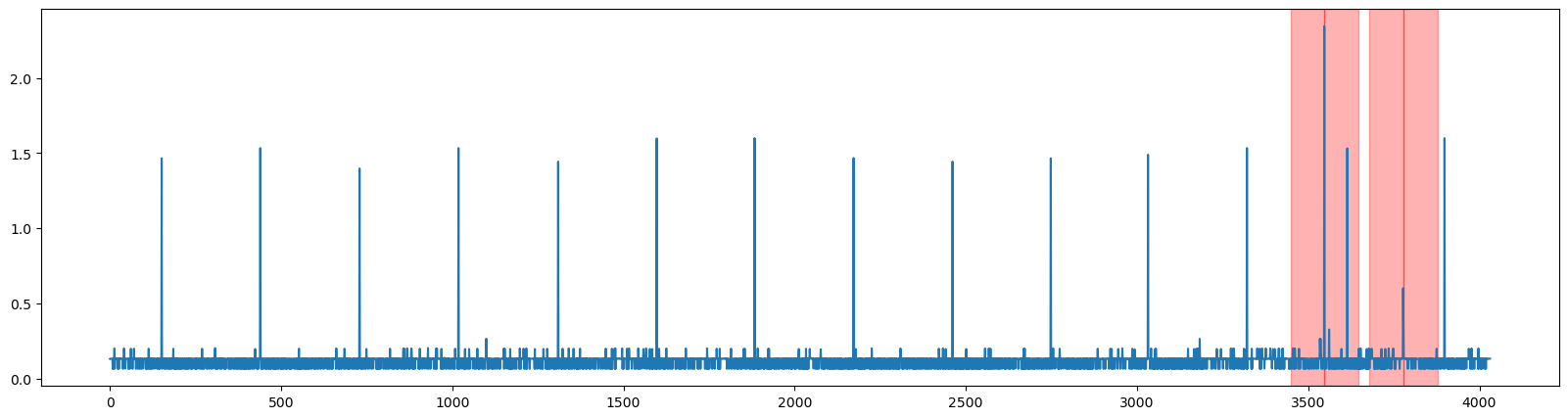}
    \caption{An example of a NAB time series with a seasonal behavior spanning over hundreds of timesteps.}
    \label{seasonal_behavior}
\end{figure*}

An important constraint of NAB-AWSCLoudWatch is that certain time series have normal seasonal behavior with a characteristic time window of hundreds of time-steps as can be seen in Fig. \ref{seasonal_behavior}. If we required our model to take them into account the SuDai encoding circuit would therefore necessitate hundreds of layers, a depth that goes way beyond what our simulation hardware can handle. Indeed the time complexity of training of a QVC of depth $k$ grows quadraticaly in $k$: one run of the circuit is linear in terms of $k$ but the number of runs to compute the gradient is proportional to the number of gates and consequently proportional to $k$.
Thus, we chose a subset of the dataset that contains all the time series where periodic behavior is smaller than the time window chosen for our implementation, which is $10$ time-steps.\\

This NAB-AWSCloudWatch dataset contains multiple time-series with different distributions, each is composed of thousands of time steps. To account for the specificities of the dataset we use a simple anomaly detection system. We implemented the proposed cQGAN using a continual learning strategy. That means the model is trained continually on a data stream.

\subsection{Continual Learning on AWSCloudWatch Data}

First, we normalize the data so that each point is contained in the interval $[0,1]$. This can be accomplished without prior knowledge of the extremous data points in the series using the information available about the monitored KPI. For example, CPU usage, if measured in watts, is bounded by the maximum hardware power usage. This is the only preprocessing performed to the series.\\ 

A pseudocode description of the anomaly detection system using continual learning with the proposed cQGAN is presented in Algorithm \ref{continual_learning}.
At each time step (line \ref{line_timestep}), we use the cQGAN to generate a data point given the contextual data in a fixed time window (line \ref{line_generate}) and compare it to the real data point. If the distance between the two is within a certain threshold $\eta$ the data point is considered normal, if not it is flagged as an anomaly (lines \ref{line_start_if}-\ref{line_end_if}). Then we train the model on the new data point by taking $N$ gradient steps. The number of steps depends on the quantum loss (lines \ref{line_start_train}-\ref{line_end_train}), which is obtained with the fidelity measurement between the generated and real data. Training is performed around anomalous points to enable the model to adapt to a potential new normal distribution that might differ from the pre-anomaly distribution. To accelerate the continual learning process, $N$ can be set to $0$ if the loss function is really close to its minimal value (smaller than some value $\epsilon$). This formalizes the idea that if the system learned the conditional distribution and encounters a new data point that falls in this distribution, it does not need to learn further and can move on to the next time-step. 

\begin{algorithm}[!h]
\SetKw{Init}{Initialize:\\}
\SetKw{Input}{Input:}
\SetKw{Output}{Output:}
\SetKw{Generate}{Generate}
\SetKwFunction{Train}{Train cQGAN}
\caption{Continual learning based anomaly detection system}\label{continual_learning}
\Init{ Time window $\Delta_t$\\
    \qquad Anomaly threshold $\eta$\\
    \qquad Gradient step policy $N$\\
    \qquad Circuit parameter vector $\boldsymbol{\theta}$ \\
    \qquad Learning rate $\alpha$}\\
\Input{Time series $\{X_t\}_{t\leq T}$}\\
\Output{Binary mapping $\sigma : \{1,...,T\} \rightarrow \{0,1\}$}\\
\For{$t \leftarrow 1$ \KwTo $T$}{ \label{line_timestep}
\tcc{Generate data and compare to detect an anomaly}
\Generate{$x_t' \leftarrow G_{\boldsymbol{\theta}}(X_{t-1},...,X_{t-\Delta_t})$}\\ \label{line_generate}
\eIf{$|x_t-x_t'|<\eta$}{ \label{line_start_if}
    $\sigma(t) \leftarrow 0$
}{$\sigma(t) \leftarrow 1$}\label{line_end_if}
\tcc{Train the cQGAN on the new data point}
\For{$i\leftarrow 1$ \KwTo $N$}{\label{line_start_train}
    \Train(\qquad \\
            \qquad $a = (X_{t-1},...,X_{t-\Delta_t})$,\\ 
            \qquad $b = X_t$,\\ 
            \qquad $\boldsymbol{\theta}$,\\
            \qquad $\alpha$\\
            )
} \label{line_end_train}
}
\end{algorithm}


\subsection{Implementation Details}
The implementation is accomplished with the Qiskit Python library~\cite{Qiskit} and its AerSimulator backend. Regarding the hyper-parameters, the learning rate is set to $\alpha =0.005$. The cQGAN considers a time window of 10 timesteps to learn and make predictions. Lastly, the number of gradient steps per data point is $N = \lfloor 10 L_{D,X} \rfloor $ unless the quantum loss is smaller than a threshold $\epsilon = 0.05$ in which case $N=0$.
In our experiments on a AMD ryzen 5 3600 cpu, the execution of the model on a single time series is typically in the order of a few hours.

\section{Results Analysis}\label{sec5}

\subsection{Raw Results}
\begin{figure*}[t!]
    \centering
    \includegraphics[scale=0.4]{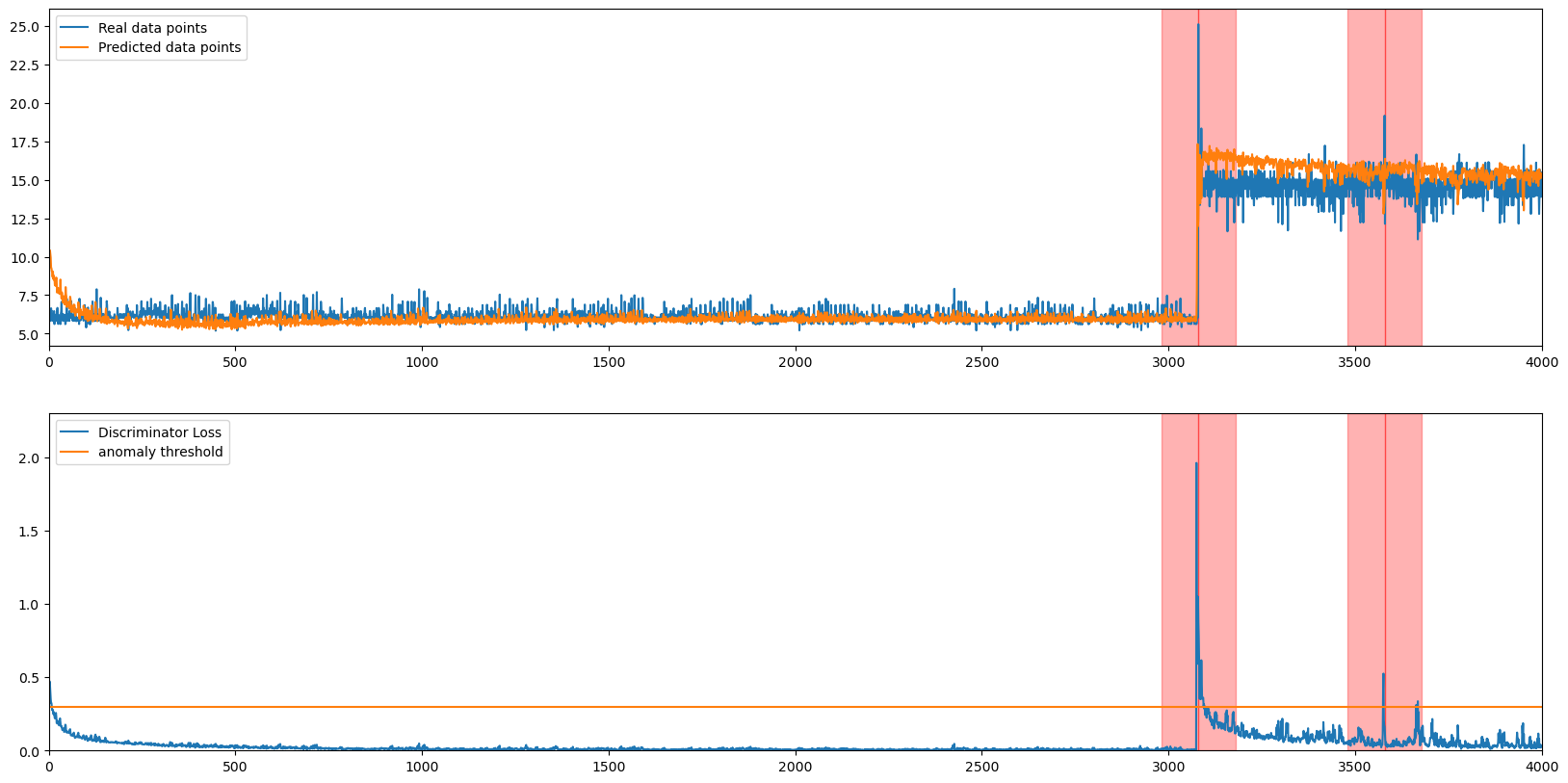}
    \caption{Example result after running the continual learning cQGAN on a time series with $\eta = 0.3$.}
    \label{example_result}
\end{figure*}

Once the algorithm has run on the entire time series we can extract the time series of predictions and the time series of quantum loss scores. For the example illustrated in Fig.(\ref{example_time_series}), we can visualize the result in Fig.(\ref{example_result}). As we can see, the system learns to replicate the signal quite closely. When an anomaly occurs in the signal, a peak shows in the quantum loss at the anomaly point. Then the model adapts to the new "normal" signal and carries on. By examining the loss series, a point is flagged as an anomaly if its associated quantum loss is higher than the threshold $\eta$.\\

A noteworthy observation in the results is that either at the initial learning phase or when an anomaly occurs, the system takes a certain time to learn enough to reduce the loss below $\eta$. This results in many subsequent points wrongly flagged as anomalies. To avoid this pitfall one strategy is to prune some of these points by  waiting for the system to adapt to the post anomaly distribution. This can be done either by waiting for a fixed number of time steps or by waiting until the loss gets smaller than a threshold that can be $\eta$ or another value. For our case we used a waiting time of 30 timesteps\\

For the example of Fig.(\ref{example_result}) and a value $\eta = 0.3$ the system detects both anomalies correctly as all values of the quantum loss above the threshold correspond to time-steps in the anomaly time windows of the time series. Two of them are the actual anomalies and the third one is an artefact of the second anomaly still contained inside the corresponding time window and is thus pruned by the NAB suggested methodology \cite{ahmad2017unsupervised}.\\

\subsection{Dynamic Threshold}

Selecting an appropriate value for $\eta$ is crucial as excessively low values for $\eta$ will lead to an excessive number of false positives  while overly high values will miss many true anomalies and will generate a large amount of false negatives. However it would not be appropriate to manually chose a value for $\eta$ that suits each time series, as this would essentially be equivalent to manually identifying anomalies. The scheme for choosing $\eta$ should be consistent across all time series. Ideally we would chose a low value for $\eta$ while avoiding to yield too many false positives. This ideal scenario is achieved when the loss reaches very low values because the model makes accurate predictions. Achieving this ideal scenario is however nearly impossible when dealing with noisy signals in one-point forecasting. Indeed, even if the model has learned a noisy data distribution and produces a noisy signal, the chance that the generated data points match the real ones is very small. So for noisy signals, the loss series will also be noisy and contain spikes as the model prediction may be quite different from reality from time to time. It is a challenge to prevent from flagging those spikes as anomalies as they are artefacts of the proposed method that is vulnerable to noise.\\

To address this challenge, we propose to use a dynamic threshold $\eta(t)$ that varies at each time step $t$. The threshold is computed as a linear model of an estimate of the noise in the loss series (Eq.~\eqref{dynamic threshold}). The idea behind is that the necessary discrepancy with respect to normality should be proportional to the noise level for identifying spikes as anomalies. Therefore, we aim to have a high threshold when the signal is noisy and a low threshold when the signal has low noise. To establish this dynamic threshold, we proceed as follows.\\

Firstly, we characterize the noise at each given time. This primary need is performed in two steps. We start by defining the local variations around a time-step $t$ as $Var(t, \delta)$ (see Eq.\eqref{variations}), where $\delta$ is the temporal distance to the furthest neighbour considered. The value of $\delta$ is typically very small. In our implementation we use $\delta = 2$ time-steps. On the other hand, the local variation estimate considers only the largest variations. The rationale behind this is that at such small timescales, prominent variations come from two sources: anomalies and noise. Indeed, a spike in the loss signifies that the model is surprised by the real data. If the model has already learned an underlying distribution, large surprises arise from elements outside of the distribution. These elements are then either noise or more significant changes in the distribution which can likely be considered as anomalies. \\

Secondly, we calculate a rough estimate of the dynamic noise level that allow excluding all the anomalies (and also some noise unfortunately). This is done through quantile calculation. We assume that anomalies are rare representing less than $0.5\%$ of the data points, and they should be located at the far end of the noise distribution. Therefore, we fix the quantile at $p=0.95$. This means we assume that $95\%$ of lower noises computed are considered as true noises. In our implementation, the noise estimator is applied within a large time window with $\tau = 500$, i.e. within 500 timestamps before the current time. The Eq.~\eqref{Noise estimator} allows calculating the estimate of the noise level. \\    

Finally, we construct the dynamic threshold $\eta$ (see Eq.~\eqref{dynamic threshold}) by starting with a base threshold $b$. Then we add the noise level estimation weighted by a parameter $a$. Ideally $b$ should be as small as possible while taking into account the $\epsilon$ parameter used earlier as $\epsilon$ represents the acceptable error for the model. We chose $b = 6 \times \epsilon = 0.3$ to be robust to small variations especially when they are close to no noise overall. The key is to select a value for $a$ that scales the noise estimation to a useful level when compared to the loss. We experimentally pick $a=2$.

\begin{equation}
    Var(t,\delta) = max\{|L_{t}-L_{i}|\}_{|i-t|\leq \delta}\label{variations}
\end{equation}
\begin{equation}
    Noise(t,\tau) = Quantile(p,\{Var(s, \delta)\}_{t-\tau< s < t}) \label{Noise estimator}
\end{equation}
\begin{equation}
    \eta(t) = a \:Noise(t, \tau) + b \label{dynamic threshold}
\end{equation}

\subsection{Experimental Results}

The results concerning the 8 time series on which we tested our model are provided in Table \ref{result_tab}.
As suggested by the NAB methodology we consider all positives within one anomaly time window to be the same and reduced to the earliest detection. We did not include the numbers of True Negatives in this table as they are not very informative in this context. Most data points are not anomalies and are not flagged. In other words, almost all the points outside of anomaly time windows are true negatives.\\

\begin{table*}[htb]
\centering
{\begin{tabular*}{40pc}{@{\extracolsep{\fill}}llllll@{}}\toprule
Time series                             &Number of  & Number of & True      & False     &False  \\
name                                    &anomalies  & positives & positives &positives  &negatives\\
\midrule
ec2\_cpu\_utilization\_5f5533           &2          &2          &2          &0          &0 \\
ec2\_cpu\_utilization\_825cc2           &2          &3          &2          &1          &0 \\
ec2\_cpu\_utilization\_ac20cd           &1          &2          &1          &1          &0 \\
ec2\_network\_in\_257a54                &1          &1          &1          &0          &0 \\
elb\_request\_count\_8c0756             &2          &3          &2          &1          &0\\
iio\_us-east-1\_i-a2eb1cd9\_NetworkIn   & 2         &1          &1          &0          &1\\
rds\_cpu\_utilization\_cc0c53           & 2         &2          &2          &0          &0\\
rds\_cpu\_utilization\_e47b3b           & 2         &2          &2          &0          &0\\
\midrule
Total                                   &14         &16         &13         &3          &1\\
\bottomrule
\end{tabular*}}{}
\caption{Results}
\label{result_tab}
\end{table*}

From the Table \ref{result_tab}, we can see there is only one false negative across all the data. It corresponds to a small anomalous spike in "iio\_us-east-1\_i-a2eb1cd9\_NetworkIn". There are a few false positives, two of them corresponding to normal (as per the labels given in NAB) but quite drastic changes in the distribution. These instances occur in datastreams "ec2\_cpu\_utilization\_825cc2" and "ec2\_cpu\_utilization\_ac20cd". In fact, both of those false positives are due to a shift back to a previous normal distribution. Since our model works on $\Delta_t =10$ time steps, it has no chance of remembering the old distribution. Therefore the model considers this change as an anomaly. The last false positive is due to noise in "elb\_request\_count\_8c0756", the most noisy stream used. The dynamic threshold allows to avoid numerous spikes caused by noise. Still, one of these spikes got flagged as an anomaly by the system as can be seen in Fig.~\ref{fullresults_2} of the Appendix. Moreover by looking at the curves for such a noisy signal, the dynamic threshold is operating at its limit with several noise spikes approaching the threshold closely.
The rest of the results can be seen in the appendix in Fig.\ref{fullresults_1} and Fig.\ref{fullresults_2}.

\section{Conclusion and Future Works}\label{sec6}

In this paper we introduced a novel method for encoding high dimensional data using a smaller number of qubits compared to the most commonly used method which is angle encoding. This method encodes high dimensional data into the quantum state while still using a linear number of gates. This allows us to process more data than with the usual angle encoding method since the number of qubits necessary needs not to scale with the number of data points at hand. The number of qubits does not depend on the size of the data but rather on the functionnality that needs to be expressed. This method also retains a complexity advantage over techniques such as amplitude encoding. In short, SuDaI solves the problem of the dimensionality of the data being encoded by shifting the limit from number of qubits to the depth of the circuit. In general, increasing the depth of a quantum circuit generates more noise which is detrimental to the performance of an algorithm. However, quantum variational algorithms have been proven to be more robust to noise both theoretically and experimentally \cite{sharma2020noise}.\\

Besides, we have applied the new encoding method to build a cQGAN for anomaly detection. For this application, we have devised a dynamic threshold that incorporates the noise level in the loss signal to enhance the precision of anomaly detection, enabling the model to handle noisy signals better. While the suggested dynamic threshold shows promising results, the challenge of noise has not been entirely eliminated. One potential solution to address this issue would be transitioning from a forecasting-based approach to a distribution-based approach, which would account for noise without requiring post-processing.\\

It is important to highlight that the experiments were conducted using network data but the model can be readily applied to anomaly detection in any kind of sequential data, provided that the time window hypothesis stands. Typical use cases might include financial time series, seismic or meteorologic data among others. Furthermore, SuDaI could be generalized for encoding generic data into a quantum state in non-conditional quantum GANs. This would enable the utilization of higher dimensional data in quantum variational applications, even within the constraints of the current limitations on the number of available qubits.

\section*{Acknowledgment}

The authors would like to thank the Natural Sciences and Engineering Research Council of Canada, Prompt, Thales and Zetane Systems, for the financial support of this research.

\ifCLASSOPTIONcaptionsoff
  \newpage
\fi



\bibliographystyle{IEEEtran}
\bibliography{IEEEabrv,Ref}

\begin{thebibliography}{10}
\providecommand{\url}[1]{#1}
\csname url@samestyle\endcsname
\providecommand{\newblock}{\relax}
\providecommand{\bibinfo}[2]{#2}
\providecommand{\BIBentrySTDinterwordspacing}{\spaceskip=0pt\relax}
\providecommand{\BIBentryALTinterwordstretchfactor}{4}
\providecommand{\BIBentryALTinterwordspacing}{\spaceskip=\fontdimen2\font plus
\BIBentryALTinterwordstretchfactor\fontdimen3\font minus \fontdimen4\font\relax}
\providecommand{\BIBforeignlanguage}[2]{{%
\expandafter\ifx\csname l@#1\endcsname\relax
\typeout{** WARNING: IEEEtran.bst: No hyphenation pattern has been}%
\typeout{** loaded for the language `#1'. Using the pattern for}%
\typeout{** the default language instead.}%
\else
\language=\csname l@#1\endcsname
\fi
#2}}
\providecommand{\BIBdecl}{\relax}
\BIBdecl

\bibitem{tsay_analysis_2005}
R.~S. Tsay, \emph{\BIBforeignlanguage{en}{Analysis of {Financial} {Time} {Series}}}.\hskip 1em plus 0.5em minus 0.4em\relax John Wiley \& Sons, Sep. 2005, google-Books-ID: ddL4tTLb\_08C.

\bibitem{craddock_analysis_1965}
\BIBentryALTinterwordspacing
J.~M. Craddock, ``The {Analysis} of {Meteorological} {Time} {Series} for {Use} in {Forecasting},'' \emph{Journal of the Royal Statistical Society. Series D (The Statistician)}, vol.~15, no.~2, pp. 167--190, 1965, publisher: [Royal Statistical Society, Wiley]. [Online]. Available: \url{https://www.jstor.org/stable/2987390}
\BIBentrySTDinterwordspacing

\bibitem{wu_network_2005}
Q.~Wu and Z.~Shao, ``Network {Anomaly} {Detection} {Using} {Time} {Series} {Analysis},'' in \emph{Joint {International} {Conference} on {Autonomic} and {Autonomous} {Systems} and {International} {Conference} on {Networking} and {Services} - (icas-isns'05)}, Oct. 2005, pp. 42--42, iSSN: 2168-1872.

\bibitem{botzen_economic_2019}
\BIBentryALTinterwordspacing
W.~J.~W. Botzen, O.~Deschenes, and M.~Sanders, ``The {Economic} {Impacts} of {Natural} {Disasters}: {A} {Review} of {Models} and {Empirical} {Studies},'' \emph{Review of Environmental Economics and Policy}, vol.~13, no.~2, pp. 167--188, Jul. 2019, publisher: The University of Chicago Press. [Online]. Available: \url{https://www.journals.uchicago.edu/doi/full/10.1093/reep/rez004}
\BIBentrySTDinterwordspacing

\bibitem{cashell2004economic}
B.~Cashell, W.~D. Jackson, M.~Jickling, and B.~Webel, ``The economic impact of cyber-attacks,'' \emph{Congressional research service documents, CRS RL32331 (Washington DC)}, vol.~2, 2004.

\bibitem{doocy2013human}
S.~Doocy, A.~Daniels, C.~Packer, A.~Dick, and T.~D. Kirsch, ``The human impact of earthquakes: a historical review of events 1980-2009 and systematic literature review,'' \emph{PLoS currents}, vol.~5, 2013.

\bibitem{tsay_time_1986}
\BIBentryALTinterwordspacing
R.~S. Tsay, ``Time {Series} {Model} {Specification} in the {Presence} of {Outliers},'' \emph{Journal of the American Statistical Association}, vol.~81, no. 393, pp. 132--141, Mar. 1986, publisher: Taylor \& Francis \_eprint: https://www.tandfonline.com/doi/pdf/10.1080/01621459.1986.10478250. [Online]. Available: \url{https://doi.org/10.1080/01621459.1986.10478250}
\BIBentrySTDinterwordspacing

\bibitem{Pang_2021}
\BIBentryALTinterwordspacing
G.~Pang, C.~Shen, L.~Cao, and A.~V.~D. Hengel, ``Deep learning for anomaly detection: A review,'' \emph{ACM Comput. Surv.}, vol.~54, no.~2, mar 2021. [Online]. Available: \url{https://doi.org/10.1145/3439950}
\BIBentrySTDinterwordspacing

\bibitem{xia2022gan}
X.~Xia, X.~Pan, N.~Li, X.~He, L.~Ma, X.~Zhang \emph{et~al.}, ``Gan-based anomaly detection: A review,'' \emph{Neurocomputing}, vol. 493, pp. 497--535, 2022.

\bibitem{schlegl_unsupervised_2017}
\BIBentryALTinterwordspacing
T.~Schlegl, P.~Seeböck, S.~M. Waldstein, U.~Schmidt-Erfurth, and G.~Langs, ``\BIBforeignlanguage{en}{Unsupervised {Anomaly} {Detection} with {Generative} {Adversarial} {Networks} to {Guide} {Marker} {Discovery}},'' Mar. 2017. [Online]. Available: \url{https://arxiv.org/abs/1703.05921v1}
\BIBentrySTDinterwordspacing

\bibitem{schlegl_f-anogan_2019}
\BIBentryALTinterwordspacing
T.~Schlegl, P.~Seeböck, S.~M. Waldstein, G.~Langs, and U.~Schmidt-Erfurth, ``\BIBforeignlanguage{en}{f-{AnoGAN}: {Fast} unsupervised anomaly detection with generative adversarial networks},'' \emph{\BIBforeignlanguage{en}{Medical Image Analysis}}, vol.~54, pp. 30--44, May 2019. [Online]. Available: \url{https://www.sciencedirect.com/science/article/pii/S1361841518302640}
\BIBentrySTDinterwordspacing

\bibitem{zenati_efficient_2019}
\BIBentryALTinterwordspacing
H.~Zenati, C.~S. Foo, B.~Lecouat, G.~Manek, and V.~R. Chandrasekhar, ``Efficient {GAN}-{Based} {Anomaly} {Detection},'' May 2019, arXiv:1802.06222 [cs, stat]. [Online]. Available: \url{http://arxiv.org/abs/1802.06222}
\BIBentrySTDinterwordspacing

\bibitem{mirza2014conditional}
M.~Mirza and S.~Osindero, ``Conditional generative adversarial nets,'' 2014.

\bibitem{de_leon_materials_2021}
\BIBentryALTinterwordspacing
N.~P. de~Leon, K.~M. Itoh, D.~Kim, K.~K. Mehta, T.~E. Northup, H.~Paik \emph{et~al.}, ``Materials challenges and opportunities for quantum computing hardware,'' \emph{Science}, vol. 372, no. 6539, p. eabb2823, Apr. 2021, publisher: American Association for the Advancement of Science. [Online]. Available: \url{https://www.science.org/doi/full/10.1126/science.abb2823}
\BIBentrySTDinterwordspacing

\bibitem{arute_quantum_2019}
\BIBentryALTinterwordspacing
F.~Arute, K.~Arya, R.~Babbush, D.~Bacon, J.~C. Bardin, R.~Barends \emph{et~al.}, ``\BIBforeignlanguage{en}{Quantum supremacy using a programmable superconducting processor},'' \emph{\BIBforeignlanguage{en}{Nature}}, vol. 574, no. 7779, pp. 505--510, Oct. 2019, number: 7779 Publisher: Nature Publishing Group. [Online]. Available: \url{https://www.nature.com/articles/s41586-019-1666-5}
\BIBentrySTDinterwordspacing

\bibitem{kim2023evidence}
Y.~Kim, A.~Eddins, S.~Anand, K.~X. Wei, E.~Van Den~Berg, S.~Rosenblatt \emph{et~al.}, ``Evidence for the utility of quantum computing before fault tolerance,'' \emph{Nature}, vol. 618, no. 7965, pp. 500--505, 2023.

\bibitem{schuld_introduction_2015}
\BIBentryALTinterwordspacing
M.~Schuld, I.~Sinayskiy, and F.~Petruccione, ``An introduction to quantum machine learning,'' \emph{Contemporary Physics}, vol.~56, no.~2, pp. 172--185, Apr. 2015, publisher: Taylor \& Francis \_eprint: https://doi.org/10.1080/00107514.2014.964942. [Online]. Available: \url{https://doi.org/10.1080/00107514.2014.964942}
\BIBentrySTDinterwordspacing

\bibitem{biamonte_quantum_2017}
\BIBentryALTinterwordspacing
J.~Biamonte, P.~Wittek, N.~Pancotti, P.~Rebentrost, N.~Wiebe, and S.~Lloyd, ``\BIBforeignlanguage{en}{Quantum machine learning},'' \emph{\BIBforeignlanguage{en}{Nature}}, vol. 549, no. 7671, pp. 195--202, Sep. 2017, number: 7671 Publisher: Nature Publishing Group. [Online]. Available: \url{https://www.nature.com/articles/nature23474}
\BIBentrySTDinterwordspacing

\bibitem{cerezo_variational_2021}
\BIBentryALTinterwordspacing
M.~Cerezo, A.~Arrasmith, R.~Babbush, S.~C. Benjamin, S.~Endo, K.~Fujii \emph{et~al.}, ``\BIBforeignlanguage{en}{Variational quantum algorithms},'' \emph{\BIBforeignlanguage{en}{Nature Reviews Physics}}, vol.~3, no.~9, pp. 625--644, Sep. 2021, number: 9 Publisher: Nature Publishing Group. [Online]. Available: \url{https://www.nature.com/articles/s42254-021-00348-9}
\BIBentrySTDinterwordspacing

\bibitem{lloyd_quantum_2018}
\BIBentryALTinterwordspacing
S.~Lloyd and C.~Weedbrook, ``Quantum {Generative} {Adversarial} {Learning},'' \emph{Physical Review Letters}, vol. 121, no.~4, p. 040502, Jul. 2018, publisher: American Physical Society. [Online]. Available: \url{https://link.aps.org/doi/10.1103/PhysRevLett.121.040502}
\BIBentrySTDinterwordspacing

\bibitem{dallaire-demers_quantum_2018}
\BIBentryALTinterwordspacing
P.-L. Dallaire-Demers and N.~Killoran, ``\BIBforeignlanguage{English}{Quantum generative adversarial networks},'' \emph{\BIBforeignlanguage{English}{Physical Review A}}, vol.~98, no.~1, p. 012324, Jul. 2018, place: College Pk Publisher: Amer Physical Soc WOS:000439409500007. [Online]. Available: \url{https://journals.aps.org/pra/abstract/10.1103/PhysRevA.98.012324}
\BIBentrySTDinterwordspacing

\bibitem{zoufal_quantum_2019}
\BIBentryALTinterwordspacing
C.~Zoufal, A.~Lucchi, and S.~Woerner, ``\BIBforeignlanguage{en}{Quantum {Generative} {Adversarial} {Networks} for learning and loading random distributions},'' \emph{\BIBforeignlanguage{en}{npj Quantum Information}}, vol.~5, no.~1, pp. 1--9, Nov. 2019, number: 1 Publisher: Nature Publishing Group. [Online]. Available: \url{https://www.nature.com/articles/s41534-019-0223-2}
\BIBentrySTDinterwordspacing

\bibitem{stein_qugan_2021}
S.~A. Stein, B.~Baheri, D.~Chen, Y.~Mao, Q.~Guan, A.~Li \emph{et~al.}, ``{QuGAN}: {A} {Quantum} {State} {Fidelity} based {Generative} {Adversarial} {Network},'' in \emph{2021 {IEEE} {International} {Conference} on {Quantum} {Computing} and {Engineering} ({QCE})}, Oct. 2021, pp. 71--81.

\bibitem{chu_iqgan_2022}
\BIBentryALTinterwordspacing
C.~Chu, G.~Skipper, M.~Swany, and F.~Chen, ``\BIBforeignlanguage{en}{{IQGAN}: {Robust} {Quantum} {Generative} {Adversarial} {Network} for {Image} {Synthesis} {On} {NISQ} {Devices}},'' Oct. 2022, arXiv:2210.16857 [quant-ph]. [Online]. Available: \url{http://arxiv.org/abs/2210.16857}
\BIBentrySTDinterwordspacing

\bibitem{niu_entangling_2022}
\BIBentryALTinterwordspacing
M.~Y. Niu, A.~Zlokapa, M.~Broughton, S.~Boixo, M.~Mohseni, V.~Smelyanskyi \emph{et~al.}, ``Entangling {Quantum} {Generative} {Adversarial} {Networks},'' \emph{Physical Review Letters}, vol. 128, no.~22, p. 220505, Jun. 2022, publisher: American Physical Society. [Online]. Available: \url{https://link.aps.org/doi/10.1103/PhysRevLett.128.220505}
\BIBentrySTDinterwordspacing

\bibitem{nielsen2001quantum}
M.~A. Nielsen and I.~L. Chuang, ``Quantum computation and quantum information,'' \emph{Phys. Today}, vol.~54, no.~2, p.~60, 2001.

\bibitem{goodfellow_generative_2014}
\BIBentryALTinterwordspacing
I.~J. Goodfellow, J.~Pouget-Abadie, M.~Mirza, B.~Xu, D.~Warde-Farley, S.~Ozair \emph{et~al.}, ``\BIBforeignlanguage{en}{Generative {Adversarial} {Networks}},'' Jun. 2014. [Online]. Available: \url{https://arxiv.org/abs/1406.2661v1}
\BIBentrySTDinterwordspacing

\bibitem{wierichs_general_2021}
\BIBentryALTinterwordspacing
D.~Wierichs, J.~Izaac, C.~Wang, and C.~Y.-Y. Lin, ``\BIBforeignlanguage{en}{General parameter-shift rules for quantum gradients},'' Jul. 2021. [Online]. Available: \url{https://arxiv.org/abs/2107.12390v3}
\BIBentrySTDinterwordspacing

\bibitem{crooks_gradients_2019}
\BIBentryALTinterwordspacing
G.~E. Crooks, ``\BIBforeignlanguage{en}{Gradients of parameterized quantum gates using the parameter-shift rule and gate decomposition},'' May 2019. [Online]. Available: \url{https://arxiv.org/abs/1905.13311v1}
\BIBentrySTDinterwordspacing

\bibitem{blazquez2021review}
A.~Bl{\'a}zquez-Garc{\'\i}a, A.~Conde, U.~Mori, and J.~A. Lozano, ``A review on outlier/anomaly detection in time series data,'' \emph{ACM Computing Surveys (CSUR)}, vol.~54, no.~3, pp. 1--33, 2021.

\bibitem{schuld2021effect}
M.~Schuld, R.~Sweke, and J.~J. Meyer, ``Effect of data encoding on the expressive power of variational quantum-machine-learning models,'' \emph{Physical Review A}, vol. 103, no.~3, p. 032430, 2021.

\bibitem{perez2020data}
A.~P{\'e}rez-Salinas, A.~Cervera-Lierta, E.~Gil-Fuster, and J.~I. Latorre, ``Data re-uploading for a universal quantum classifier,'' \emph{Quantum}, vol.~4, p. 226, 2020.

\bibitem{sim_expressibility_2019}
\BIBentryALTinterwordspacing
S.~Sim, P.~D. Johnson, and A.~Aspuru-Guzik, ``\BIBforeignlanguage{en}{Expressibility and entangling capability of parameterized quantum circuits for hybrid quantum-classical algorithms},'' \emph{\BIBforeignlanguage{en}{Advanced Quantum Technologies}}, vol.~2, no.~12, p. 1900070, Dec. 2019, arXiv:1905.10876 [quant-ph]. [Online]. Available: \url{http://arxiv.org/abs/1905.10876}
\BIBentrySTDinterwordspacing

\bibitem{ahmad2017unsupervised}
S.~Ahmad, A.~Lavin, S.~Purdy, and Z.~Agha, ``Unsupervised real-time anomaly detection for streaming data,'' \emph{Neurocomputing}, vol. 262, pp. 134--147, 2017.

\bibitem{Qiskit}
{Qiskit contributors}, ``Qiskit: An open-source framework for quantum computing,'' 2023.

\bibitem{sharma2020noise}
K.~Sharma, S.~Khatri, M.~Cerezo, and P.~J. Coles, ``Noise resilience of variational quantum compiling,'' \emph{New Journal of Physics}, vol.~22, no.~4, p. 043006, 2020.

\end{thebibliography}
%

\appendix[Full Results]\label{appendix}

\begin{figure*}[h]
\centering
    \includegraphics[scale=0.6]{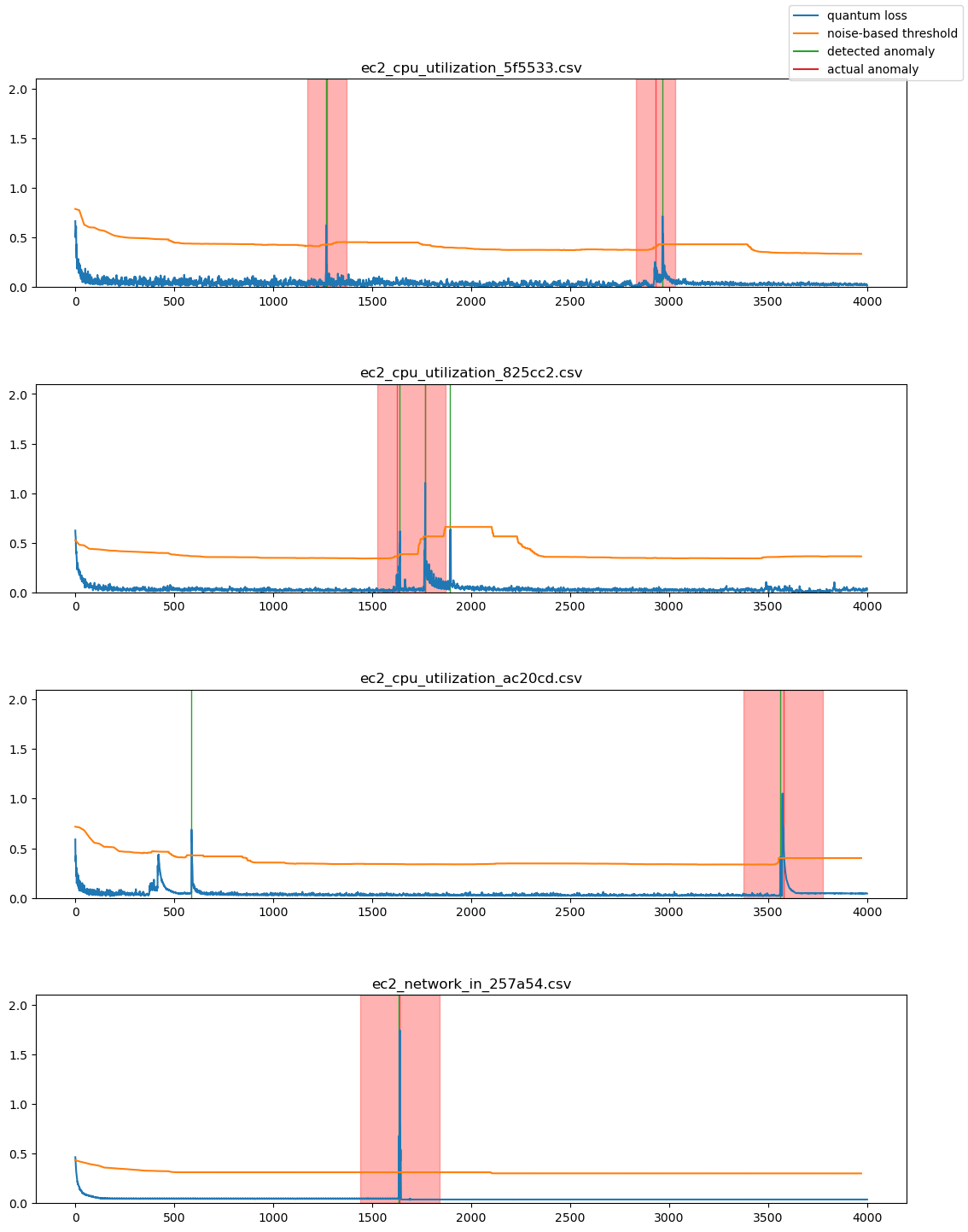}
    \caption{Full results for all selected time series part 1}
    \label{fullresults_1}
\end{figure*}
\begin{figure*}[h]
\centering
    \includegraphics[scale=0.6]{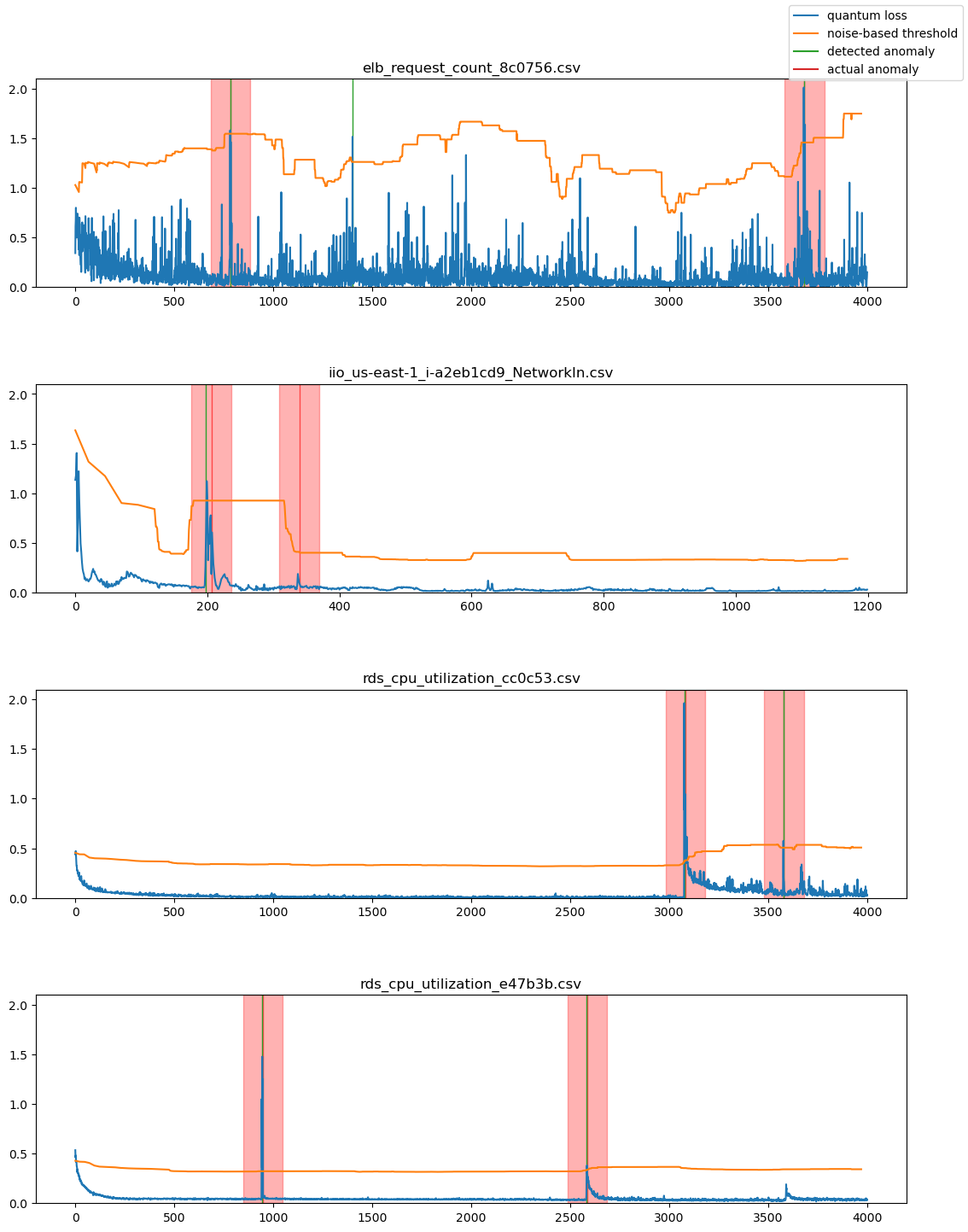}
    \caption{Full results for all selected time series part 2}
    \label{fullresults_2}
\end{figure*}

\end{document}